\documentclass[]{article} 
\usepackage[preprint]{colm2026_conference}

\usepackage{microtype}
\usepackage{hyperref}
\usepackage{url}
\usepackage{booktabs}
\usepackage{float}

\usepackage{lineno}

\definecolor{ourpurple}{HTML}{7760FB}
\definecolor{ourgreen}{HTML}{5CAD70}

\definecolor{darkblue}{rgb}{0, 0, 0.5}
\hypersetup{colorlinks=true, citecolor=ourgreen, linkcolor=ourgreen, urlcolor=ourgreen}

\usepackage{amsmath}
\usepackage[noabbrev]{cleveref}
\usepackage{caption}
\usepackage{subcaption}
\usepackage{graphicx}
\usepackage{inconsolata}
\usepackage{underscore}
\usepackage{tabularray}
\usepackage{siunitx}

\usepackage{titlesec}
\titlespacing*{\paragraph}{\parindent}{0.25ex}{1ex}
\titlespacing*{\section}{0pt}{3pt}{3pt}
\titlespacing*{\subsection}{0pt}{3pt}{3pt}

\usepackage{enumitem}
\setlist[enumerate,itemize]{topsep=0pt,itemsep=0pt,leftmargin=18pt}

\newcommand{\ann}[1]{\mbox{\texttt{#1}}}
\newcommand{\behavior}[1]{\mbox{\texttt{#1}}}

\newcommand{\failtag}[1]{\mbox{\texttt{#1}}}

\newcommand{\arch}[1]{{\color{ourpurple}{#1}}}
\newcommand{\stat}[1]{#1}

\title{A paradox of AI fluency}

\author{Christopher Potts$^{1,2}$ and Moritz Sudhof$^{1}$ \\
${}^{1}$Bigspin AI, ${}^{2}$Stanford University\\
\texttt{\{cgpotts, moritz\}@bigspin.ai} \\
}

\begin{document}

\ifcolmsubmission
\linenumbers
\fi

\maketitle

\begin{abstract}
How much does a user's skill with AI shape what AI actually delivers for them? This question is critical for users, AI product builders, and society at large, but it remains underexplored. Using a richly annotated sample of \stat{27K} transcripts from WildChat-4.8M, we show that fluent users take on more complex tasks than novices and adopt a fundamentally different interactional mode: they iterate collaboratively with the AI, refining goals and critically assessing outputs, whereas novices take a passive stance. These differences lead to a paradox of AI fluency: fluent users experience more failures than novices -- but their failures tend to be visible (a direct consequence of their engagement), they are more likely to lead to partial recovery, and they occur alongside greater success on complex tasks. Novices, by contrast, more often experience invisible failures: conversations that appear to end successfully but in fact miss the mark. Taken together, these results reframe what success with AI depends on. Individuals should adopt a stance of active engagement rather than passive acceptance. AI product builders should recognize that they are designing not just model behavior but user behavior; encouraging deep engagement, rather than friction-free experiences, will lead to more success overall.\footnote{Code/data: \url{https://github.com/bigspinai/bigspin-fluency-outcomes}}
\end{abstract}

\section{Introduction}

How does expertise with AI translate into success with AI tools and products? This is an increasingly critical question for users, AI product builders, educators, and societies at large, but it remains underexplored.  In the present paper, we seek to fill this gap in our understanding. Our analysis is based in the \mbox{WildChat-4.8M} dataset of human--AI interactions. We sampled \stat{1K} transcripts per month from \stat{May 2023 to July 2025 (27 months)}, and we richly annotated these transcripts with information about visible and invisible failures \citep{potts2026invisible}, task complexity, and the user's level of AI fluency.

Three key sets of findings emerge from this analysis: 
\begin{enumerate}
\item \textbf{Ambition}: Fluent users take on more complex tasks than novice users do. On our 5-point complexity scale, there is a \stat{1.6-point gap} between the mean task complexity for the highest fluency users and the lowest fluency ones (\stat{3.1 vs.~1.5}; see \cref{fig:complexity}).

\item \textbf{Interactional mode}: Fluent users adopt an \emph{augmentative} stance: they iterate collaboratively with the AI, refining their goals and critically assessing outputs as they go \citep{anthropic2026fluency}.  By contrast, novice users are \emph{delegative}: they tend to passively accept the AI's plans and responses, and often fail to get the information they need as a result. This contrast is stark: \stat{93\%} of high-fluency user interactions are labeled augmentative in our data vs.~\stat{under 1\%} for the lowest fluency users (\cref{fig:style}).

\item \textbf{A fluency paradox}: Fluent users experience \emph{more} failures than novices: \stat{64\%} of their transcripts contain at least one failure indicator vs.~\stat{24\%} for the lowest fluency group (\cref{fig:failures}).  However, \stat{59\%} of the high-fluency users' failures are \emph{visible} rather than invisible in the sense of \citealt{potts2026invisible}. This is a direct consequence of their augmentative mode: experts steer the AI when it goes off track. In contrast, just \stat{12\%} of the lowest fluency users' failures are visible. Additionally, fluent users are more likely to achieve a partial recovery (\cref{fig:archetypes}), to take on complex tasks, and to succeed at complex tasks (\cref{fig:complexity}).
\end{enumerate}

These findings show that AI expertise predicts whether AI is a lottery or a tool. This leads to concrete practical advice for individual users, educators, and AI system designers. For example, individuals should adopt a stance of active engagement rather than passive acceptance; this runs counter to the prominent narrative of AI as all-knowing superintelligence, so it is critical that educators reinforce this message. At the same time, AI system designers should develop interfaces and AI interaction patterns that encourage deep engagement, rather than friction-free experiences. It is well-known that even expert users can fall into a mode of uncritical acceptance of AI responses \citep{lee2025critical,ryser2025calibrated}, so these efforts should target all users. This product development approach does risk making users less satisfied in the moment, but our findings indicate that it will lead to greater success for them long-term. Taken together, our results show that success with AI depends on more than model capabilities. User engagement is also a deciding factor.

\section{Related work}

\paragraph{AI fluency}
A number of recent papers offer general frameworks for AI literacy (\citealt{jin2024glat,liu2025literacy,li2025gfactor}; see \citealt{Jones_2026} for an overview). \citet{rosala2026ailiteracy} provides an accessible  overview of concepts related to AI fluency, with a focus on prompting and critically reviewing model outputs. \citet{yan2026literacy} similarly assess users on their knowledge of risks ranging from misinformation to security. Our own approach to AI fluency is inspired by \citet{anthropic2026fluency}, itself based on research by \citealt{dakan2025framework}. The \citeauthor{anthropic2026fluency} study finds that ``the most common expression of AI fluency is augmentative -- treating AI as a thought partner, rather than delegating work entirely'' (see also \citealt{jin2025agency}). We provide what is in effect a conceptual replication of this result using \mbox{WildChat-4.8M}, but our key contribution lies in connecting fluency with outcomes. 

\paragraph{AI expertise and outcomes}
Agency is a prominent theme of recent work on how AI expertise relates to outcomes \citep{li2025gfactor,jin2025agency,prather2024widening}, and it is central to  \citet{anthropic2026fluency} and our own work as well. \citet{Brynjolfsson2025} and \citet{cui2026genai} identify connections between domain expertise and productivity. These works find that novices see larger productivity gains from AI than experts. These results, however, are not directly about expertise with AI, but rather relate to experience in the domain in which AI has been deployed. Our results, by contrast, are specifically focused on users' expertise with AI, and we do not measure their expertise in the task domains in which they are operating.

\paragraph{AIs as (unusual) social actors}
One often-overlooked aspect of expertise with AI is acquiring an intuitive understanding of AIs as social actors. People tend to default to assuming that, because these systems use language so fluently, they will be pragmatic, human-like conversationalists \citep{Grice75}  with a familiar sense of how common ground is established and negotiated \citep{Clark:Schreuder:1983}. Experts quickly learn that this is not true; AIs' fluency tends to mask their fragmentary understanding of the context \citep{fried-etal-2023-pragmatics,shao2025collaborative} and of the world (\citealt{vafa2024worlds,vafa2025foundationmodelfoundusing,mancoridis2025potemkin}; for alternative perspectives, see  \citealt{gurnee2024,li2024emergentworld,tsvilodub_emergent_2026}). The specific augmentative behaviors that the current paper identifies in high-fluency users can be seen as indirect efforts to manage the unusual interactional nature of AI. 

\paragraph{The future of work}
Discussions of AI fluency invite questions about how AI is likely to affect employment for workers in different industries and at different skill levels \citep{eloundou2024,brynjolfsson2025canaries,Brynjolfsson2025,machovec2025incorporating,shao2026future,wang2025ai,hazra2025aisafety,louie2026upksill,weilnhammer2026}. Our work complements these efforts by showing how expert behaviors relate to success with AI in open-ended, largely unstructured contexts like the one provided by ChatGPT. 

\section{Data}

Our dataset is derived from \mbox{WildChat-4.8M},\footnote{\url{https://huggingface.co/datasets/allenai/WildChat-4.8M}} which is an extension of \mbox{WildChat-1M} \citep{zhao2024wildchat}. For both WildChat variants, anonymous users were given free access to ChatGPT in exchange for having their deidentified transcripts collected and shared publicly. We use the non-toxic subset of WildChat-4.8M, which has a total of 3,199,860 conversations from the period April 9, 2023, to August 1, 2025. The oldest LLM in the dataset is \stat{GPT-3.5-turbo} and the newest is \stat{GPT-4.1-mini}. (See \cref{appendix:models} for additional details on the distribution of models.)

\subsection{Annotated subset}

For our annotations and subsequent analyses, we randomly sampled 1K English-language examples per month from \stat{May 2023 to July 2025} (inclusive). This resulted in a dataset of \stat{27K} examples. We experienced a small amount of random data loss during our annotation runs; in the end, our dataset has \stat{26,958} annotated examples, with each month having \stat{between 993 and 1,000} examples.

\subsection{Two exogenous events: Midjourney prompting and Blockman Go}\label{sec:exogenous}

In the course of our analysis of \mbox{WildChat-4.8M}, we discovered two large, highly specific subsets of the data that have a significant impact on \mbox{WildChat-4.8M} as a whole, in addition to shaping our findings. We call these subsets ``Midjourney'' and ``Blockman''. We use ``Standard'' for the dataset with all these examples removed, and ``Full'' for our entire dataset.

The ``Midjourney'' transcripts all involve variations of a Midjourney prompting strategy that went viral in 2023.\footnote{It is difficult to trace the origins of this strategy precisely, but this Github repository is a likely origin or early distribution point: \url{https://github.com/friuns2/BlackFriday-GPTs-Prompts/blob/main/gpts/midjourney-ai.md}} It seems to have been largely unsuccessful with the models of the time, in that the prompts generated seem often not to match the user's specification. There are \stat{4,335} such transcripts in our dataset.
Essentially all of these (\stat{98\%}) are marked as \stat{\ann{moderate}} fluency users by our annotator, which drives up the rate of \ann{moderate} considerably, and \stat{81\%} of them have task complexity \stat{3}, which similarly shapes the complexity distribution.

The ``Blockman'' transcripts are identified by the line ``You are an AI named Zexy - and are currently chatting in a Blockman Go group''. Subsequent turns seem to involve distinct users, though it is difficult to say for sure. Our current best guess is that the \mbox{WildChat-4.8M} service was used opportunistically by someone running an online agent. We estimate that there are over 200K of these ``Blockman'' transcripts in WildChat-4.8M. In our annotated sample, there are \stat{1,654} such cases. Almost all of them (\stat{89\%}) have task complexity \stat{1}, and \stat{64\%} are marked as \stat{minimal} fluency.

Which dataset is to be preferred, ``Full'', or subsets that exclude ``Blockman'' and/or ``Midjourney''? We think there is no single answer to this question. On the one hand, both ``Blockman'' and ``Midjourney'' represent real usage patterns that AI services are likely to experience. On the other hand, they are exogenous events, and future events might have a different character.

In what follows, we generally use the ``Standard'' dataset -- that is, we remove all ``Blockman'' and ``Midjourney'' examples from our analyses. This dataset has \stat{20,969} cases. Our goal is to be sure we are not inadvertently providing a picture just of these unusual subsets. Were we in the role of product developers, though, these transcripts would be of significant interest.

\section{Methods}

\subsection{User fluency annotations}\label{sec:fluencyann}

We annotated all of the transcripts in our sample using a fluency protocol inspired by \citealt{anthropic2026fluency}. The full annotation script, which includes the entire prompt, is included in the code release for this report, in \stat{\texttt{tag_user_fluency.py}}. The annotations assigned to each transcript have the following structure: 
\begin{enumerate}
  \item \textbf{Transcript Summary:} One sentence describing what the user was trying to accomplish.

  \item \textbf{Interaction Style:} \texttt{augmentative} $|$ \texttt{delegative} $|$ \texttt{other}

  \item \textbf{Fluency Behaviors:} A list of zero or more \ann{behavior\_name} annotations with metadata:
  \begin{enumerate}
    \item \ann{behavior\_name}: One of \stat{17} fluency behavior categories.
    \begin{itemize}
      \item \textbf{Strength:} 1 $|$ 2 $|$ 3, with custom definitions for each \textit{behavior\_name}.
      \item \textbf{Evidence:} Brief quote or paraphrase (under 30 words).
      \item \textbf{Turn:} Where the behavior is manifested in the transcript.
      \item \textbf{Notes:} Optional context.
    \end{itemize}
  \end{enumerate}

  \item \textbf{Anti-Fluency Behaviors:} A list of zero or more \textit{behavior\_name} annotations with metadata:
  \begin{enumerate}
    \item \textit{behavior\_name}: One of \stat{7} anti-fluency behavior categories.
    \begin{itemize}
      \item \textbf{Evidence:} Brief quote or paraphrase (under 30 words).
      \item \textbf{Notes:} Optional context.
    \end{itemize}
  \end{enumerate}


  \item \textbf{Fluency Assessment:} \texttt{high} $|$ \texttt{moderate} $|$ \texttt{low} $|$ \texttt{minimal}

  \item \textbf{Assessment Rationale:} 2--3 sentences explaining the overall fluency judgment, citing the most important behaviors observed.
\end{enumerate}
We used \mbox{Sonnet 4.5} (\texttt{claude-sonnet-4-5-20250929}) as our fluency annotation model. In this report, we concentrate on the interaction style annotations, the fluency behaviors, and anti-fluency behaviors.

\subsection{Task complexity annotations}\label{sec:complexityann}

We also annotated our sample for information about the complexity of the user's task. These annotations are based on the full transcript, including all available \mbox{WildChat-4.8M} metadata, and they were done by \mbox{Sonnet 4.5} (\texttt{claude-sonnet-4-5-20250929}). The primary values we use from this annotation phase are the \ann{confidence_score} values, which are on a 1--5 scale. The annotations also include labels for %
\stat{%
    \ann{cognitive_complexity}, %
    \ann{domain_expertise}, %
    \ann{scope_ambiguity}, %
    \ann{novelty}, and %
    \ann{task_type}%
}. The annotation script is included in our code release as \stat{\texttt{tag_task_complexity.py}}.

\subsection{Failure mode annotations}\label{sec:failann}

Finally, we annotated our sample using the two-stage annotation approach of \citealt{potts2026invisible}.\footnote{\url{https://github.com/bigspinai/bigspin-invisible-failure-archetypes}} In stage~1, three LLMs %
(\mbox{Claude Sonnet 4.6} (\texttt{claude-sonnet-4-6}), 
\mbox{Claude Opus 4.6} (\texttt{claude-opus-4-6}), and %
\mbox{GPT-5.4} (\texttt{gpt-5.4-2026-03-05})) %
separately annotate each transcript with a large number of quality signals, using an annotation protocol that these LLMs developed themselves over multiple rounds, with guidance from the authors. In stage~2, a final annotator LLM (\mbox{Claude Opus 4.6}) uses these quality signals to infer two kinds of annotation:
\begin{enumerate}
\item \textbf{Basic failure mode classification:} \failtag{visible}, \failtag{invisible}, or \failtag{mixed} if a failure indicator is detected somewhere in the transcript; \failtag{none} if no failure is detected.
\item \textbf{Invisible failure archetypes:} zero or more of the invisible failure archetypes of \citealt{potts2026invisible} if the basic failure mode is \failtag{invisible} or \failtag{mixed}. If the failure is \failtag{visible}, then only \arch{Visible failure} is assigned. If the failure mode is \failtag{none}, then no archetypes are assigned.
\end{enumerate}
The set of archetypes is 
\stat{%
    \arch{The confidence trap},    
    \arch{The silent mismatch}, 
    \arch{The drift},
    \arch{The death spiral},
    \arch{The contradiction unravel}, 
    \arch{The walkaway},
    \arch{The partial recovery},
    and
    \arch{The mystery failure}}. %
For convenience, we repeat the definitions of these archetypes from \citealt{potts2026invisible} in \cref{appendix:archdefs}. 

For transcripts created through \stat{October 2023}, we subsample transcripts, with their failure annotations, from \citealt{potts2026invisible} and add the fluency annotations described in \cref{sec:fluencyann}. For transcripts starting in \stat{November 2023}, we randomly sample English-language cases from WildChat-4.8M and annotate them using the same annotation scripts and protocols as in \citealt{potts2026invisible} (in addition to adding our fluency annotations).

The overall distribution of archetypes for our dataset is given in \cref{appendix:archdist}; it qualitatively matches the distribution for the much larger but older sample of transcripts used by \citealt{potts2026invisible}.

\section{The AI fluency landscape}

The present section seeks to provide a picture of AI fluency in our dataset. Our goal is to lay the foundation for studying how fluency and failure modes relate to each other (\cref{sec:failures}).

\subsection{AI fluency distribution}

\begin{figure}[tp]
    \centering
    \includegraphics[width=0.8\linewidth]{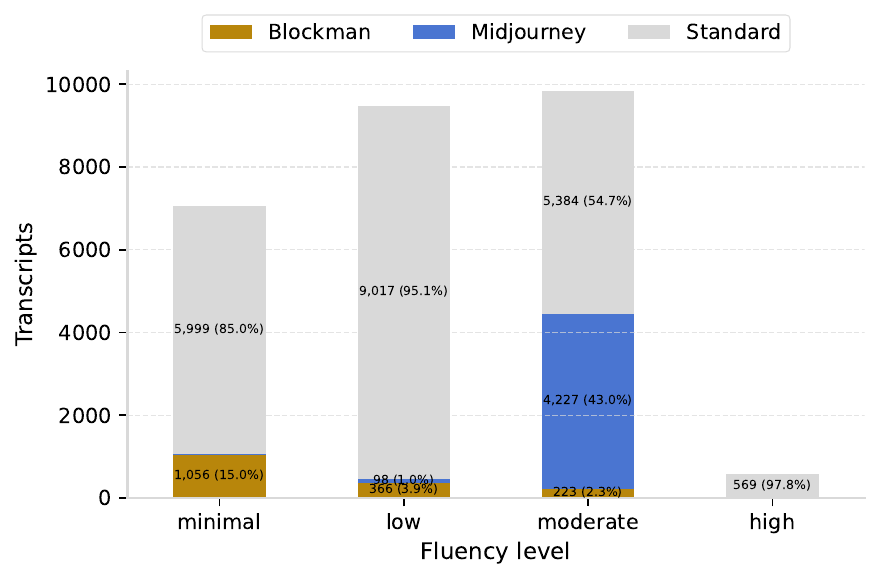}
    \caption{Overall fluency distribution. For descriptions of the three variants of the dataset, ``Standard'', ``Midjourney'', and ``Blockman'', see \cref{sec:exogenous}.}
    \label{fig:fluency-dist}
\end{figure}
 
\begin{figure}[tp]
    \includegraphics[width=1\linewidth]{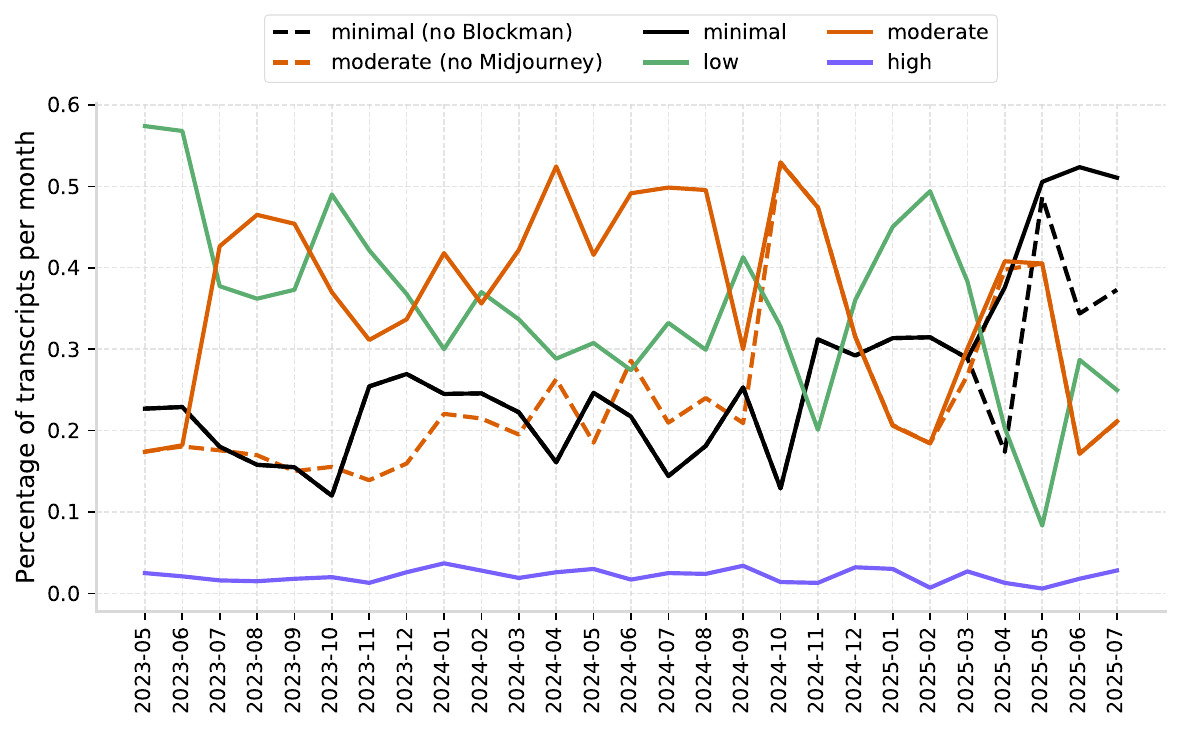}
    \caption{Fluency levels over time. The solid lines represent the full dataset. The \stat{dashed orange} line is for the dataset without the ``Midjourney'' examples (all of which appear \stat{before October 2024}, and the \stat{dashed black} line is for the dataset with the ``Blockman'' examples removed (all appear \stat{after March 2025}).}
    \label{fig:fluency-time}
\end{figure}

\Cref{fig:fluency-dist} shows the distribution of fluency levels across our entire dataset. Here, we have divided the dataset into the three groups described in \cref{sec:exogenous}. This makes it evident that including ``Midjourney'' cases boosts the rate of \ann{moderate} fluency users, and including ``Blockman'' cases very substantially boosts the rate of \ann{minimal} fluency users. Both these changes are highly consequential for the complexity and failure analyses we report later, so we largely rely on the ``Standard'' subset. Nonetheless, the other subsets represent real and systematic usage patterns that are relevant to system developers.

For all three subsets of the dataset (and for the dataset as a whole), the rate of high-fluency users is low. This may reflect the nature of the WildChat service, though the rapid growth of AI in general \citep{chatterji2025chatgpt,views4you2025aitools}  probably means that most services are dominated by low-fluency users.

\subsection{Fluency over time}

\Cref{fig:fluency-time} shows the distribution of fluency levels over time. For this plot, we group transcripts by month and track their relative proportions. This reveals a few noteworthy patterns. First, the rate of high-fluency users has remained steady (and very low) throughout the timespan of the dataset. Second, the rate of low-fluency users is trending downward. Third, the rate of minimal fluency users is trending upward. 

These trends are affected by ``Midjourney'' and ``Blockman''. The dashed lines track the effects of removing these datasets. We show just the \ann{moderate} line after ``Midjourney'' removal and the \ann{minimal} line after ``Blockman'' removal, to avoid further cluttering the display. The other trend lines are affected in only negligible ways because they contain few or no examples from these special groups.

The effects of the ``Midjourney'' examples are strong: removing them significantly lowers the rate of \ann{moderate} users in early parts of the data (where all these examples are located). In the full dataset, there seems to be a downward trend for these users, but the ``Midjourney'' examples drive this trend. (The ``Midjourney'' prompts effectively disappear after \stat{September 2024}, so the dashed and solid \stat{orange} lines are nearly identical from that point onward.)

The ``Blockman'' effects are more modest and constrained. These examples occur exclusively in the period beginning in \stat{April--July 2025}, and they create a sharp upward trend in minimal fluency users in that period. When we remove all these transcripts, we get the more modest (and noisier) upward trend indicated by the dashed line. (Prior to this period, the two lines are identical because there are no earlier ``Blockman'' transcripts.) 

As above, it is perhaps not clear which of these trend lines to emphasize when thinking about the details of these usage patterns. However, for any perspective we take, it seems clear that the largest areas of user growth are in the lower-fluency categories. This is significant for developers of AI systems because, as we will see, these users need the most help when it comes to truly succeeding with AI.

\subsection{AI fluency and interactional style}\label{sec:behaviors}

\begin{figure}[tp]
    \centering
    \includegraphics[width=1\linewidth]{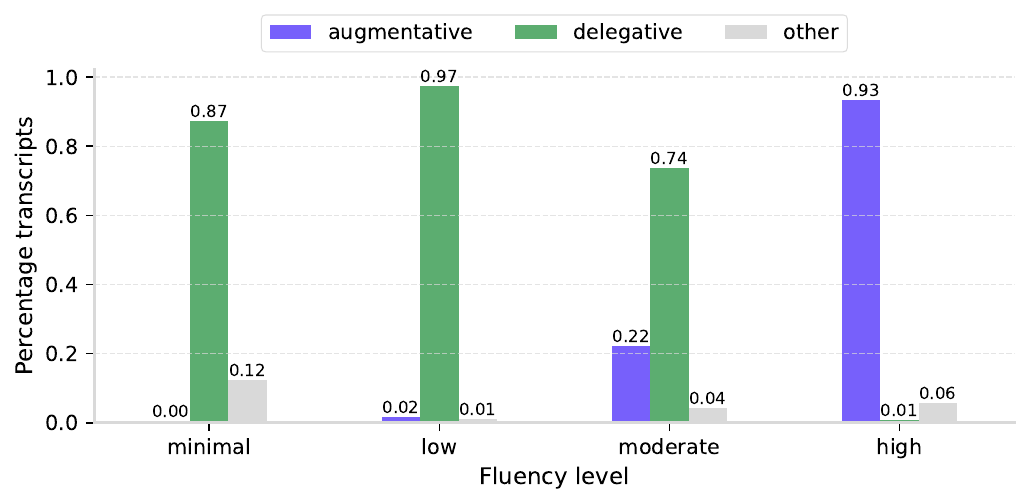}
    \caption{Fluency and interactional style for the ``Standard'' dataset (the three subsets defined in \cref{sec:exogenous} look essentially the same). Only high-fluency users consistently display the augmentative style that correlates with task success.}
    \label{fig:style}
\end{figure}

Our annotation protocol (\cref{sec:fluencyann}) includes a three-way category for interaction style: \texttt{augmentative} and \texttt{delegative}, and a catch-all category \texttt{other}. The relationship between these categories and fluency is summarized in \cref{fig:style}. This distribution is not meaningfully affected by the ``Midjourney'' or ``Blockman'' subsets, but we present results for the ``Standard'' dataset for simplicity.

These results strongly echo those reported by \citealt{anthropic2026fluency} and reflect an issue that every AI system designer should consider: high-fluency users negotiate with the AI and refine its outputs until they get what they need, whereas low-fluency users acquiesce. Since success with present-day AI virtually requires an augmentative approach for all but the simplest tasks, we can forecast from this plot alone that low-fluency users will experience less success with AI than high-fluency users.

\begin{figure}[tp]
    \begin{subfigure}[b]{0.48\linewidth}
    \centering
    \includegraphics[width=1\linewidth]{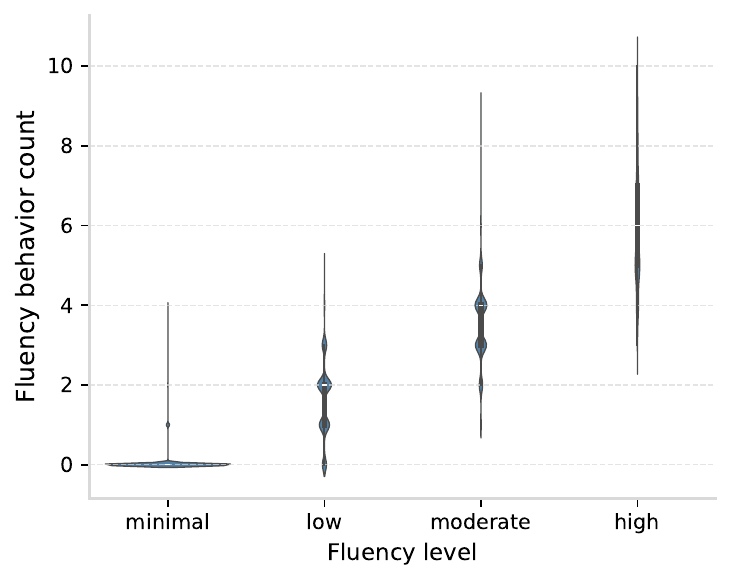}
    \caption{Fluency behaviors.}
    \end{subfigure}
    \hfill 
    \begin{subfigure}[b]{0.48\linewidth}
    \centering
    \includegraphics[width=1\linewidth]{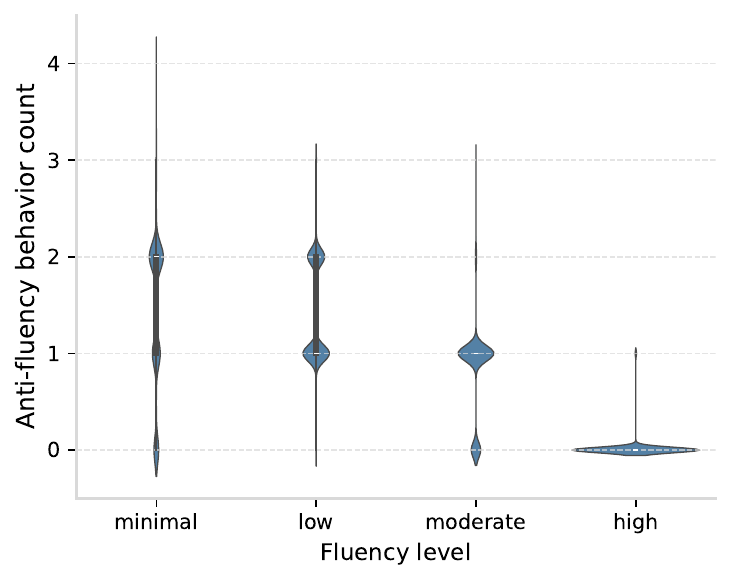}
    \caption{Anti-fluency behaviors.}
    \end{subfigure}
    \caption{Mean rates of fluency and anti-fluency behavior counts across fluency levels, in the ``Standard'' dataset.}
    \label{fig:behaviors}
\end{figure}

\Cref{fig:behaviors} deepens the pattern from \cref{fig:style}. Here, we study the average rate of the fluency and anti-fluency behavior tags assigned by our annotator (\cref{sec:fluencyann}), again using our ``Standard'' dataset. The trends are again  stark: high-fluency users show many high-fluency behaviors, and low-fluency users show many anti-fluency behaviors.

\begin{figure}[tp]
    \centering
    \includegraphics[width=1\linewidth]{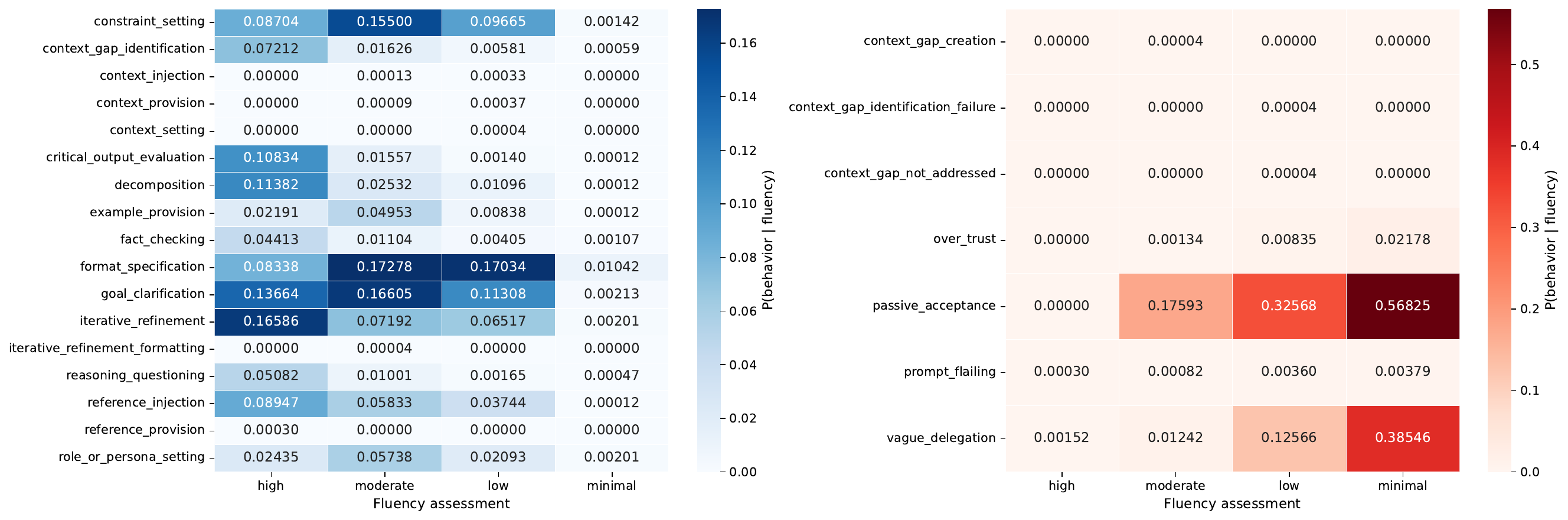}
    \caption{Fluency and anti-fluency behaviors by fluency level in the ``Standard'' dataset. For these visualizations, we defined the distributions $P(\ann{behavior} \mid \ann{fluency})$ for each fluency level, defined over the union of all fluency and anti-fluency behaviors, and then we segmented the distribution by these two behavior groups. The fluency behaviors are in the blue heatmap and the anti-fluency behaviors are in the red heatmap.}
    \label{fig:behavior-dist}
\end{figure}

\Cref{fig:behavior-dist} deepens the picture. Here, we show the combined distribution of fluency and anti-fluency behaviors across fluency categories (again in the ``Standard'' dataset). The fluency behaviors are in the blue section and the anti-fluency ones are in the red section. It is immediately evident that high-fluency users make much more use of fluency behaviors and, in turn, display very few anti-fluency behaviors. The converse is true for low-fluency users. One pattern that stands out is that of \behavior{iterative_refinement}: it is highly characteristic of high-fluency users and essentially absent from low-fluency transcripts. A counterpart of this is \behavior{passive_acceptance}, which is the strongest single indicator of minimal- and low-fluency. 

\section{AI fluency and AI failures}\label{sec:failures}

In the previous section, we provided an overview of the landscape of AI fluency in our data. We now move to studying how fluency relates to users' variable level of success with AI. We first study the overall distribution of failures and their relative visibility, and then we look at the interactions between fluency and the invisible failure archetypes of \citealt{potts2026invisible}.

\subsection{Failure rates across fluency levels}\label{sec:failure-fluency}

\begin{figure}[tp]
    \centering
    \includegraphics[width=1\linewidth]{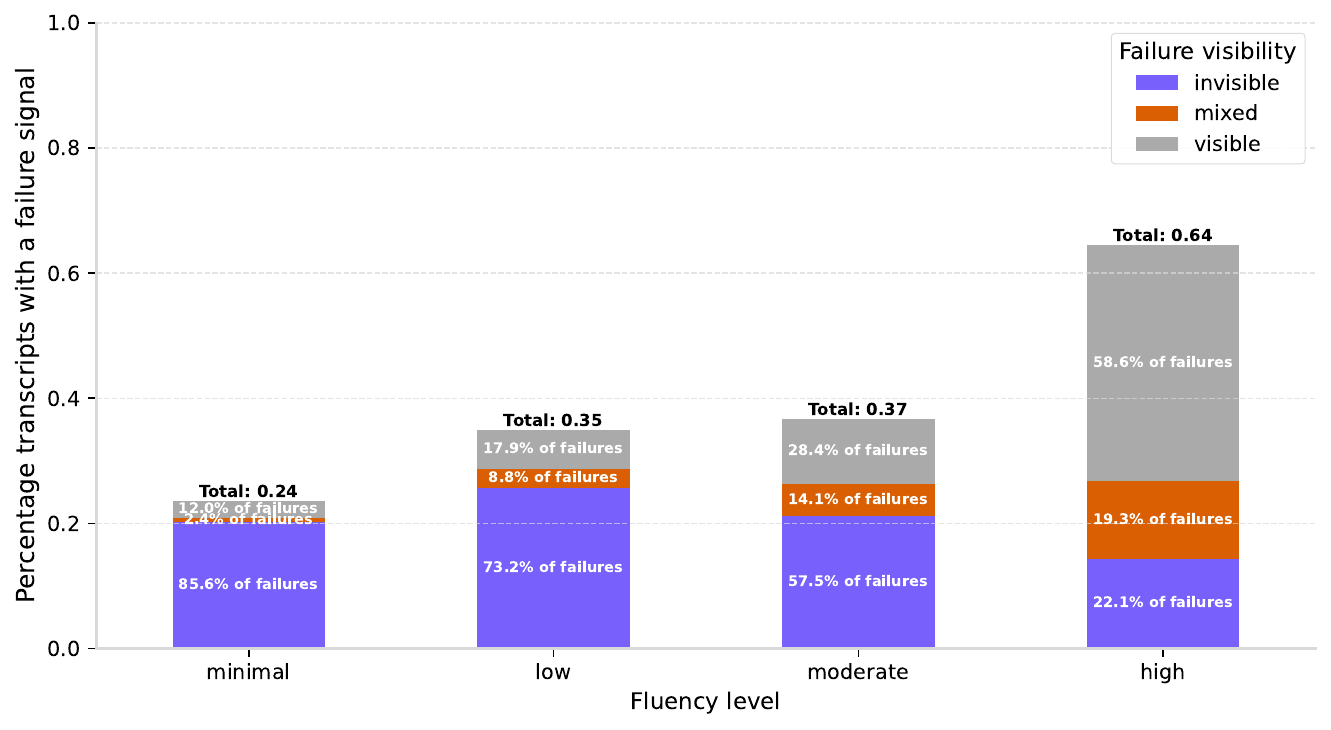}
    \caption{Failure rates across fluency levels, for the ``Standard'' variant of the dataset (\cref{sec:exogenous}). The percentage of failures for each fluency level are given as annotations on the bar segments. Failure rates rise consistently by fluency level, but \emph{invisible} failure rates fall by fluency level. Our assessment of this is that high-fluency users adopt an augmentative mode in which they respond visibly to failures, whereas low-fluency users are less likely to give an overt indication of failure. \Cref{fig:complexity} makes the connection with task complexity.}
    \label{fig:failures}
\end{figure}

\Cref{fig:failures} tracks failure rates across fluency levels, in the ``Standard'' dataset. There are two striking trends in this plot. 
First, as users gain experience with AI, their rate of invisible failures drops. For the \ann{minimal} category, \stat{85.6\%} of their failures are invisible, whereas the corresponding rate for the \ann{high} fluency group is \stat{22.1\%}.
Second, as fluency rises, failure rates overall rise as well, from \stat{24\%} for the \ann{minimal} group to \stat{64\%} for the \ann{high} group.  This is initially counterintuitive, until we see that the rise in failures is accounted for by a rise in \emph{visible} failures. Both of these trends are undoubtedly related to the fluency behaviors that we studied in \cref{sec:behaviors}: as users gain more expertise with AI, they are more inclined to push back, to refine iteratively with the AI, and to respond to failures in how the AI is behaving. In aggregate, this more interactional mode leads to more visible failures. However, as we will see in the next section, the picture is overall more positive for high-fluency users, even though their failure rates are significantly higher.

\subsection{The relationship between fluency, task complexity, and task success}\label{sec:complexity}

\begin{figure}[tp]
    \centering
    \includegraphics[width=1\linewidth]{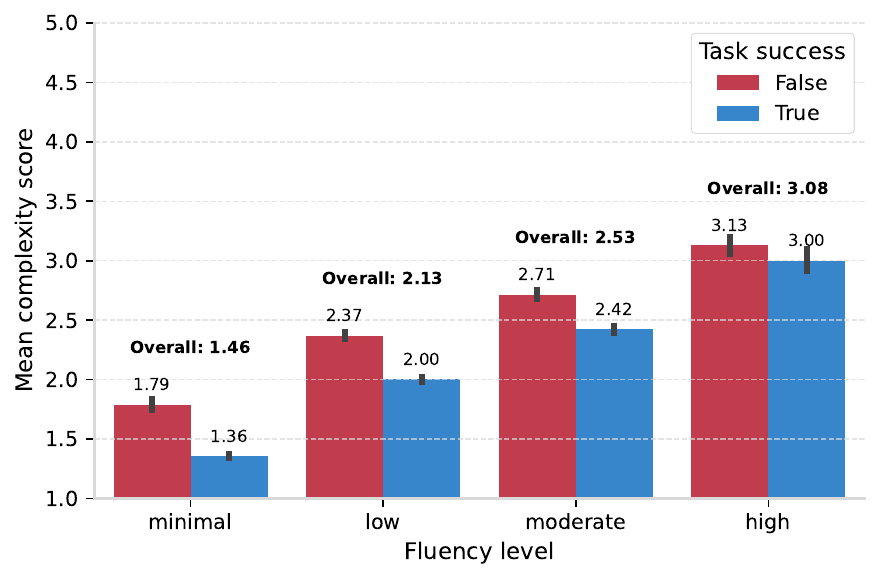}
    \caption{Fluency, task complexity, and success. Fluency is strongly correlated with increased task complexity, for both successful and unsuccessful interactions.}
    \label{fig:complexity}
\end{figure}

\Cref{fig:complexity} summarizes the overall relationship between fluency, task complexity, and task success. The plot is organized by fluency level, and the y-axis gives the mean task complexity according to our \ann{complexity_score} annotations. We have further broken down the subgroups by failure and success (defined as the absence of any failure signals from our annotation process; \cref{sec:failann}).

This picture helps us understand why high-fluency users experience more failure overall, as we saw in \cref{fig:failures}: high-fluency users take on much more complex tasks than low-fluency users. Moreover, they are more successful at these tasks, and we venture that this is \emph{because} they adopt a more augmentative approach to AI, responding to failures and generally steering the AI towards better outcomes even in challenging scenarios.

\subsection{Regression models}\label{sec:lme}

We have presented evidence that increased fluency correlates with increased visibility of AI failures (\cref{sec:failure-fluency}) and increased task success (\cref{sec:complexity}). However, a number of other factors certainly shape these outcomes as well. For example, we have seen that complexity is a significant effect (\cref{fig:complexity}). We also expect longer conversations to correlate with fluency and create opportunities for augmentative behavioral responses (\cref{sec:behaviors}). And the task domain is also likely to be related in complex ways to all of these variables.

This raises the question of whether fluency itself has explanatory power in predicting failure visibility and success. To address this, we developed two generalized linear mixed-effects models, one with \ann{success} as the dependent variable (defined over the ``Standard'' dataset) and one with \ann{visible_failure} as the dependent variable (defined over the subset of failure cases in ``Standard''). Both  encode \ann{fluency} as a numerical value (1--4; 4 = \ann{high}). The other fixed-effects predictors are \ann{n_turns}, \ann{complexity_score}, and \ann{fluency_behavior_count}. To account for clustering within domains, we included a random intercept for each domain. Overall, this is the maximal mixed-effect model structure that converged for our datasets \citep{Barr:Levy:Scheepers:Tily:2013}. The full details of these models are given in \cref{appendix:lme}. 

Overall, the results are very intuitive and support our claims that fluency is a predictor of success and of failure visibility. In both fitted models, the coefficient for \ann{fluency} is positive (fluency contributes to success and to failure visibility) and significant ($p < 0.01$).

In the \ann{success} model, the coefficients for \ann{n_turns} and \ann{complexity_score} are significant ($p < 0.001$) and negative, and the coefficient for \ann{fluency_behavior_count} is negative but not significant. This indicates that all these factors reduce success rates, as one would expect. \ann{fluency} has a (significant) positive coefficient, indicating that higher fluency leads to higher success, all things considered. 

In the \ann{visible_failure} model, all the predictors are significant ($p < 0.001$) except \ann{fluency_behavior_count}. \ann{n_turns} and \ann{fluency_behavior_count} are positive, as we would expect: visibility increases rapidly with conversation length and is shaped by fluency behaviors. The \ann{complexity_score} coefficient is negative -- complexity reduces failure visibility.

Overall, these models confirm our more descriptive results and strongly support the claim that higher fluency is a significant predictor of success and of increased failure visibility.

\subsection{Invisible failure archetypes across fluency levels}\label{sec:archetypes}

\begin{figure}[tp]
    \centering
    \includegraphics[width=1\linewidth]{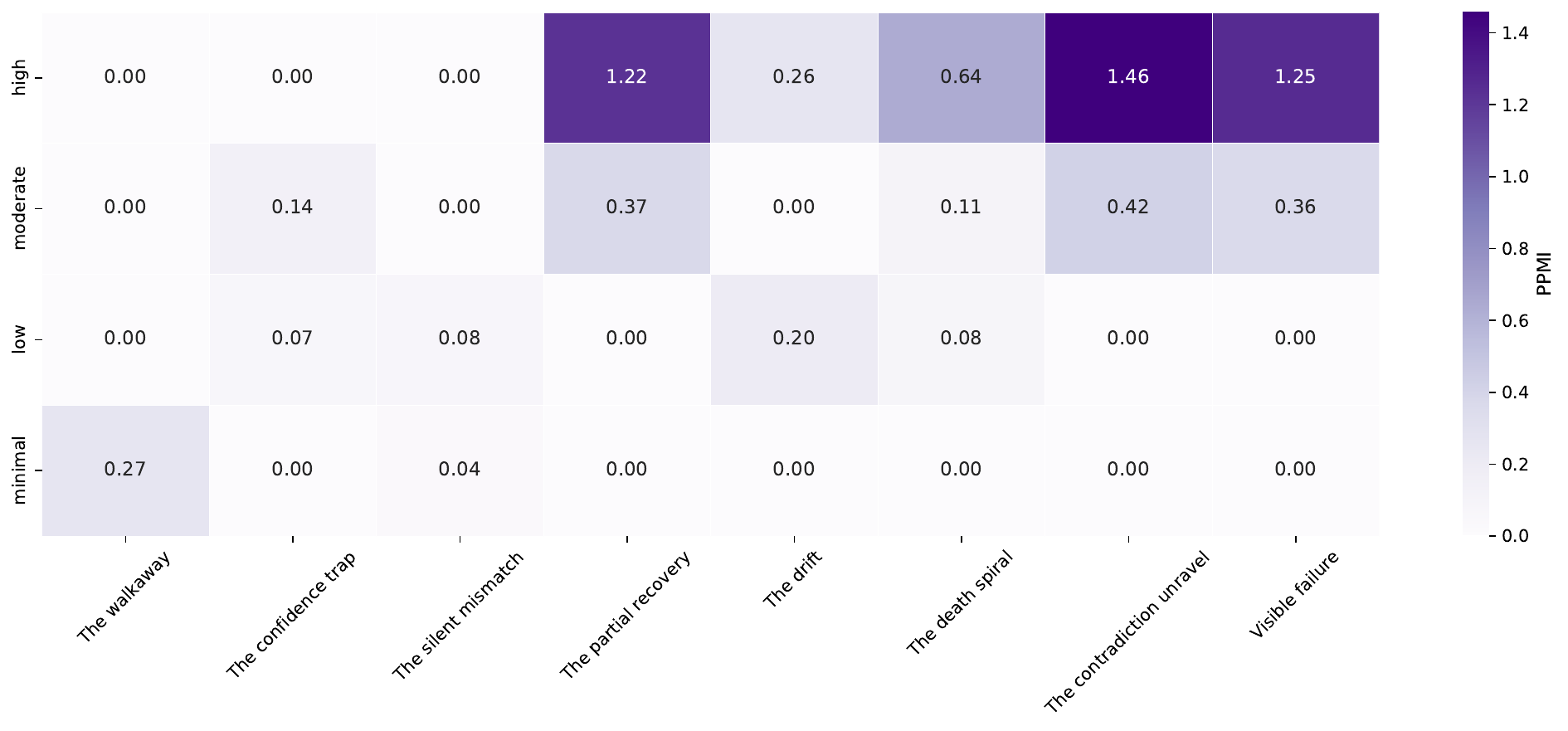}
    \caption{Invisible failure archetypes across fluency levels. The heatmap is a PPMI matrix. The noteworthy trend is that high-fluency users are most strongly associated with the \arch{Visible failure} (as we saw in \cref{fig:failures}). Additionally, while they are also associated with \arch{The drift}, \arch{The death spiral}, and \arch{The contradiction unravel}, they also stand out when it comes to \arch{The partial recovery}. By contrast, the lowest fluency group is associated with \arch{The walkaway}.}
    \label{fig:archetypes}
\end{figure}

We now take advantage of the invisible failure archetype annotations that are included in our dataset. As discussed in \cref{sec:fluencyann}, this annotation protocol is taken from \citealt{potts2026invisible}. Any transcript marked as having an invisible or mixed failure can be assigned one or more of these invisible failure archetypes. 

\Cref{fig:archetypes} provides a heatmap of the relationship between the archetypes and the fluency levels.  To create this matrix, we first compiled a co-occurrence matrix between the archetypes and the fluency levels, and then we re-weighted that matrix using positive pointwise information (PPMI; \citealt{church-hanks-1990-word,Bullinaria2007}):
\[
\text{PPMI}(X, a_{i}, b_{j}) = \max\left(0, \frac{P(X_{ij})}{P(X_{i*})\cdot P(X_{*j})}\right)
\]
Here $X$ is the matrix of co-occurrences between archetypes and fluency levels, $P(X_{ij})$ is the probability of archetype $a_{i}$ and fluency level $b_{j}$ occurring together, and $P(X_{*i})$ and $P(X_{*j})$ are row and column probabilities, respectively.

The goal of this PPMI-reweighting is to account for the different frequencies of both the archetypes and the fluency levels, and to identify co-occurrence patterns that are larger than we would expect if the archetype and fluency variables were independent of each other. 

The PPMI-matrix tells a clear story. High-fluency users have a strong association with \arch{Visible failure}, as we saw in \cref{sec:failure-fluency}, but also strong associations with \arch{The partial recovery}, \arch{The death spiral},   \arch{The contradiction unravel} and \arch{The drift}. One could summarize this by saying that high-fluency users encounter contradictions and off-topic drift, but they find ways to recover from these events. Thus, although their failure rates are higher overall, they are also more likely to iterate with the AI toward some kind of success than any other fluency level. 

Overall, this picture reinforces our core claim that high-fluency users succeed with AI, not because they enter into a passive role, but rather because they know how to work with the AI and manage its weaknesses and strengths in order to achieve success. 

\section{Discussion and conclusion}

Our overarching question is what characterizes expert use of AI products in the present moment and how these behaviors relate to patterns of success and failure with AI. In a data-driven analysis of a sample of WildChat-4.8M transcripts, we find that experts operate in an augmentative mode in which they work iteratively with the AI to refine goals and plans, check responses, and steer the interaction towards success. Less expert users tend to be more passive and accepting by comparison (\cref{sec:behaviors}). These findings complement the prior work by \citet{anthropic2026fluency}, which arrived at similar conclusions.

We go further and connect these behaviors with patterns of success. We find that, paradoxically, high-fluency users experience higher failure rates (\cref{sec:failure-fluency}). However, these failures stem from experts' augmentative mode, which makes failure more visible and leads to a higher likelihood of recovery (\cref{sec:archetypes}). As a result, high-fluency users are able to successfully tackle harder problems with AI (\cref{sec:complexity}).

We think these findings have profound implications for AI product development and AI education. On the product development side, engineers and designers should be aware of how users' different fluency levels will affect how they interact with systems. Experts may push back naturally (although it is well documented that even they can enter into a passive mode), whereas low-fluency users will require encouragement to take these actions. 

Fundamentally, these results show that there is a persistent gap between what AI can do and what AI actually delivers for most users. The gap exists because AI outcomes depend not just on the AI system's capabilities, but also on the user's ability to work effectively. When things go wrong -- which they often do, often invisibly -- it's generally the user, not the model, who determines whether the failure gets caught, whether the conversation recovers, and whether the interaction ends in something useful or in a subtle misfire that nobody notices.

When something impressive happens, we credit the model. When something fails, we blame the model. However, the people actually getting value from AI systems today are a significant factor here, and they are doing something specific and learnable: they push back, they iterate, they question outputs, they refine goals mid-conversation. It is critical that we help users acquire these skills, and critical that we develop AI products and services in ways that encourage these complex interactions, otherwise even the best models of the future will fail to deliver for most users.

\bibliography{colm2026_conference}
\bibliographystyle{colm2026_conference}

\newpage
\appendix

\crefalias{section}{appendix}

\section*{Supplementary materials}

\section{WildChat model distribution}\label{appendix:models}

\begin{table}[H]
    \centering
\begin{tabular}{lr}
\toprule
 & count \\
clean_wildchat_model &  \\
\midrule
gpt-3.5-turbo & 10,695 \\
gpt-4o-mini & 4,371 \\
gpt-4o & 4,193 \\
gpt-4.1-mini & 3,797 \\
gpt-4 & 3,493 \\
o1-preview & 265 \\
o1-mini & 161 \\
\bottomrule
\end{tabular}
    \caption{Distribution of models in our sample of WildChat-4.8M.}
    \label{tab:models}
\end{table}

\section{Invisible failure archetype definitions}\label{appendix:archdefs}

The following are the invisible failure archetype definitions from \citealt{potts2026invisible}.

\paragraph{The Confidence Trap} The AI presents incorrect information with unwarranted certainty, and the user accepts it without challenge. The danger is that the interaction looks successful — the AI sounds authoritative, the user seems satisfied — but the user walks away with wrong information. Look for: factual errors delivered without hedging, fabricated specifics, user building on incorrect premises.

\paragraph{The Silent Mismatch} The AI addresses a different goal than the user intended, but the response is plausible enough that neither party flags the disconnect. The AI "answers a question the user didn't ask." Look for: response that is competent but off-target, user's actual need going unaddressed, subtle misinterpretation of the request.

\paragraph{The Drift} The conversation gradually or abruptly loses its connection to the user's original goal. The AI may elaborate on tangentially related topics, add unrequested content, or respond to its own interpretation rather than the user's intent. Can be gradual (verbosity creeping off-topic over turns) or sudden (AI addresses a different but related topic). Look for: response relevance declining over turns, AI addressing adjacent but wrong goals, user's specific requirements getting lost.

\paragraph{The Death Spiral} The conversation enters a repetitive loop. The user keeps asking or correcting, and the AI keeps producing the same or similar output without incorporating the feedback. No progress despite continued effort. Look for: repeated similar responses across turns, user corrections that don't result in changes, escalating user frustration with static AI behavior.

\paragraph{The Contradiction Unravel} The AI contradicts its own prior statements without acknowledging the change. In earlier turns it said X; now it says not-X, with no "I was wrong" or "on reflection." This can erode the user's ability to determine what's actually correct. Look for: incompatible claims across turns, unstated reversals, user potentially confused about which version to trust.

\paragraph{The Walkaway} The conversation ends without resolution and without the user explicitly signaling failure. The user simply stops engaging. This is the hardest archetype to identify because the absence of a signal IS the signal. Look for: unresolved user goal, final AI response that doesn't fully address the need, no subsequent user message, conversation ending without natural closure.

\paragraph{The Partial Recovery} The conversation hits a clear failure but partially recovers. The AI or user identifies the problem and course-corrects, but the recovery is incomplete — the user gets some value but the original goal is not fully met. Look for: error followed by correction, improvement in response quality across turns, but remaining gaps or unaddressed aspects.

\paragraph{The Mystery Failure} The user's goal was not achieved, but no specific failure pattern from the above list explains why. The conversation just... didn't work, and it's hard to point to a specific breakdown. This is a catch-all that flags gaps in our analytical framework. Use sparingly — prefer a specific archetype if one fits even partially.

\section{Overall archetype distribution}\label{appendix:archdist}

\begin{figure}[H]
    \centering
    \includegraphics[width=1\linewidth]{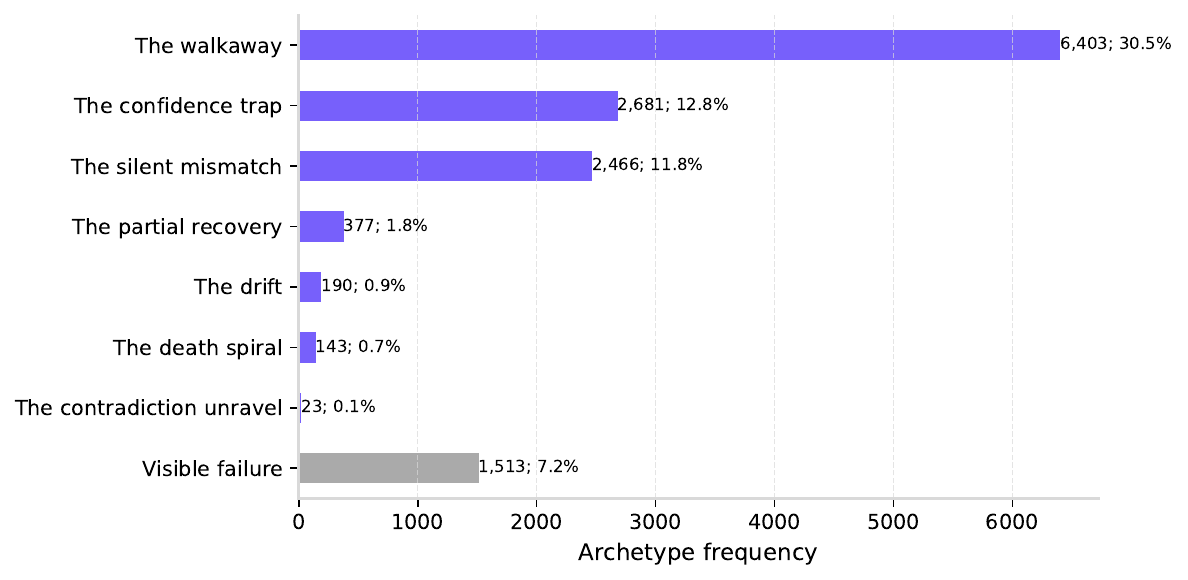}
    \caption{Archetype distribution for our dataset.}
    \label{fig:archdist}
\end{figure}

\section{Additional regression model details}\label{appendix:lme}

As discussed in \cref{sec:lme}, we fit a generalized linear mixed-effects model to better understand the extent to which fluency predicts task success and failure visibility.

\subsection{Success model}

Our success model is specified as follows using R/lmer notation \citep{bates2015lme4}:
\begin{verbatim}
glmer(
  is_success ~ 1 + fluency_scalar + n_turns + complexity_score + 
  fluency_behavior_count +  (1 | domains),
  data=model_df,
  family=binomial,
  control=glmerControl(optimizer="bobyqa"))    
\end{verbatim}
Here, \ann{model_df} is the ``Standard'' dataset. (Qualitatively similar results are obtained for ``Full''.) \ann{is_success} is defined in terms of the absence of any failure indicators in our annotations (\cref{sec:failann} and \cref{sec:failures}). \ann{fluency_scalar} encodes our four-way fluency categorization as a numerical value (1--4; 4 = \ann{high}).
\ann{complexity_score} is given directly in terms of those annotations (1--5; see \cref{sec:complexityann} and \cref{sec:complexity}), and \ann{fluency_behavior_count} is the number of fluency tags (\cref{sec:fluencyann} and \cref{sec:behaviors}). The term \verb"1 | domains)" is a a random intercept for each domain. Domains with 100 or more transcripts are kept as-is, and the rest are clustered into a single \ann{other} category, which ends up covering \stat{2\%} of the examples. The model was fit in R (version 4.5.3) using the \texttt{lme4} package (2.0.1) with the BOBYQA optimizer.

The following summarizes the resulting fitted model:

\begin{table}[H]
\centering
\begin{tabular}{l r r l}
\toprule
& coefficient & s.e. &  \\
\midrule
(Intercept) & \num{2.087} & \num{0.118} & *** \\
\ann{fluency} & \num{0.111} & \num{0.041} & ** \\
\ann{n_turns} & \num{-0.236} & \num{0.009} & *** \\
\ann{complexity_score} & \num{-0.517} & \num{0.021} & *** \\
\ann{fluency_behavior_count} & \num{-0.005} & \num{0.020} & \\
\midrule
Random intercept standard deviation & \multicolumn{3}{c}{\num{0.475}}  \\
Number of observations & \multicolumn{3}{c}{20,969} \\
R2 Marg. & \multicolumn{3}{c}{\num{0.176}} \\
R2 Cond. & \multicolumn{3}{c}{\num{0.229}} \\
RMSE & \multicolumn{3}{c}{\num{0.43}} \\
\bottomrule
\end{tabular}
\caption{Success model summary. **$: p < 0.01$; ***$: p < 0.001$}
\end{table}

\subsection{Failure visibility model}

Our failure visibility model follows the same structure as our success model, but the model is fit to the subset of cases with a failure indicator (\ann{fail_df}) and the dependent variable is the visibility of the failure:
\begin{verbatim}
glmer(
  is_visible ~ 1 + fluency_scalar + n_turns + complexity_score + 
  fluency_behavior_count +  (1 | domains),
  data=fail_df,
  family=binomial,
  control=glmerControl(optimizer="bobyqa"))    
\end{verbatim}
The following summarizes the resulting fitted model:

\begin{table}[H]
\centering
\begin{tabular}{l r r l}
\toprule
& coefficient & s.e. &  \\
\midrule
(Intercept) & \num{-2.376} & \num{0.198} & *** \\
\ann{fluency} & \num{0.691} & \num{0.075} & *** \\
\ann{n_turns} & \num{0.131} & \num{0.009} & *** \\
\ann{complexity_score} & \num{-0.463} & \num{0.045} & *** \\
\ann{fluency_behavior_count} & \num{0.071} & \num{0.037} &  \\
\midrule
Random intercept standard deviation &  \multicolumn{3}{c}{\num{0.680}} \\
Number of observations &  \multicolumn{3}{c}{6,917} \\
R2 Marg. &  \multicolumn{3}{c}{\num{0.193}} \\
R2 Cond. &  \multicolumn{3}{c}{\num{0.293}} \\
RMSE &  \multicolumn{3}{c}{\num{0.38}} \\
\bottomrule
\end{tabular}
\caption{Failure visibility model summary. ***$: p < 0.001$}
\end{table}

\end{document}